\documentclass{article}

\usepackage{Parxiv}

\usepackage[utf8]{inputenc} 
\usepackage[T1]{fontenc}    
\usepackage{hyperref}       
\usepackage{url}            
\usepackage{booktabs}       
\usepackage{amsfonts}       
\usepackage{nicefrac}       
\usepackage{microtype}      
\usepackage{lipsum}
\usepackage{fancyhdr}       
\usepackage{graphicx}       
\graphicspath{{media/}}     

\usepackage{amsmath}
\usepackage{amssymb} 
\usepackage{graphicx}

\usepackage{algorithm}
\usepackage{algpseudocode}

\usepackage{tabularx}

\title{DGPExplainer: Generating Prototype Sequence Explanations for Dynamic Graph Neural Networks
}

\author{
  Jinlong Hu, Shixu Fan, Chongzhang Li \\
   School of Computer Science and Engineering\\
  South China University of Technology, Guangzhou, China.\\
  \texttt{jlhu@scut.edu.cn} \\
}

\begin{document}
\maketitle

\begin{abstract}
Graph Neural Networks (GNNs) have been successfully applied to learning from graph-structured data. However, applying them to dynamic graphs—where nodes and edges evolve over time—presents significant challenges, particularly regarding explainability. While explanation techniques for static GNNs are well established, they do not readily extend to dynamic GNNs (DGNNs) due to the added complexity introduced by temporal dynamics and spatiotemporal dependencies, especially when generating model-level explanations. To address this gap, we introduce DGPExplainer, a post-hoc, prototype-based explainer specifically designed for DGNN models that transforms their complex decision-making processes into human-understandable, analogical, case-based explanations. DGPExplainer first employs local explanation methods to identify important elements within temporal snapshots of dynamic graph sequences. It then decomposes the graph at each time step into distinct prototypes, thereby symbolizing the dynamic graph sequence. Finally, we design a sequence generation method that uses these symbolized sequences to learn and generate class-specific dynamic graph pattern sequences that serve as prototype explanations. Experiments on both synthetic and real-world datasets demonstrate that our method effectively produces prototype-based explanations for DGNN models across multiple classes.
\end{abstract}


\section{Introduction}
In recent years, Graph Neural Networks (GNNs) have achieved significant success in learning from graph-structured data across a wide range of fields, including social network analysis, molecular structure characterization, biochemistry and drug discovery, fraud detection, and recommendation systems. In practical applications, the nodes and edges within graph-structured data often evolve over time, necessitating the development of dynamic GNNs (DGNNs) to effectively model these temporal changes, such as dynamic brain network analysis and communication system modeling \cite{1,2,3}. Despite their effectiveness, these models frequently operate as opaque systems, making their decision-making processes difficult to interpret. This lack of transparency undermines trust, accountability, and the ability to identify or mitigate biases, thereby posing significant challenges to their deployment in real-world settings \cite{4, 7}.

In response to the opacity of the deep models, Explainable Artificial Intelligence (XAI) has attracted significant research interest. In this context, explanation methods for static GNNs have been extensively studied \cite{201}, including instance-level (local) explanations that reveal the reasoning behind the model’s decisions for specific samples, and model-level (global) explanations that characterize the general behavior of the GNN model. However, techniques that perform well on static graphs are not easily applicable to dynamic graph models due to the additional complexity introduced by temporal dynamics and spatiotemporal dependencies. For example, dynamic graph models complicate the derivation of comprehensive model-level summaries for global explanations—a task that is considerably more straightforward in static graph settings. In static GNNs, model-level explanations typically involve identifying a compact set of critical nodes, edges, or motifs \cite{25}. In contrast, for DGNNs, the relevance of specific structural or temporal patterns often depends on contextual factors \cite{202}. For instance, a triangular motif may be considered significant only if it persists for at least five consecutive time steps. Capturing such conditional importance in a global representation presents substantial challenges.

Case-Based Reasoning (CBR) \cite{203} posits that human knowledge acquired through experience is stored as specific examples, and that individuals solve new problems by recalling and adapting similar past cases rather than relying on abstract rules. This case-oriented cognitive mechanism provides a natural foundation for XAI. When explaining predictions from black-box models, prototype explanations employ CBR’s similarity-based reasoning: they select representative prototype samples from each class in the training data and explain the model’s decisions by comparing input features with those prototypes \cite{204}. By presenting a set of easily understandable examples—such as typical local patterns like the black-and-white stripes on zebras—prototype explanations help users intuitively understand how the model makes decisions.

In this study, we focus on post-hoc prototype explanations for DGNN models. We propose a Dynamic Graph Prototype Explainer, called DGPExplainer, which generates representative dynamic graph examples to elucidate the behavior of trained DGNN models. Initially, DGPExplainer employs local explanation methods to extract key elements from sequences of dynamic graphs. It then decomposes the graph at each time point into distinct prototypes, thereby symbolizing the dynamic graph sequence. Finally, we design a sequence generation algorithm that uses these symbolized sequences to learn and generate prototype sequences that explain the model within their respective classes. Experimental results on both synthetic and real-world datasets demonstrate that DGPExplainer effectively provides prototype-based explanations for dynamic graph neural network models across various classes.


\section{Related Work}

\subsection{Dynamic graph neural network}

Dynamic Graph Neural Networks are generally categorized into two main types  \cite{205}. The first type addresses continuous-time dynamic graphs, which track the graph's changes continuously by modeling its evolution over time. In this approach, graphs are typically represented using time series data combined with event timestamps. The second type involves discrete-time dynamic graphs, where the graph is captured at fixed, regular intervals, producing a complete snapshot of the graph at each discrete time point. Consequently, these graphs are naturally represented as a sequence of static graph snapshots. In this paper, we focus on the second type—discrete-time dynamic graph neural networks. 

\subsection{Explanation of dynamic graph neural network}

Explanation methods for static GNN models have been extensively explored, including instance-level approaches such as GNNExplainer\cite{206}, GraphLIME \cite{207} and GCN-LRP \cite{208}, as well as model-level explanations of static graphs, such as XGNN \cite{25}, GNNInterpreter \cite{26}, GLGExplainer \cite{13} and DGX\cite{209}.

Some studies have focused on explaining dynamic GNN models. For example, DGExplainer\cite{19} proposes a method of correlation backpropagation, which redistributes the output activation scores of dynamic GNN to the correlations of the neurons in the previous layer, and iterates in reverse until obtaining the correlation scores of the input nodes. He\cite{20} et al. improved the static graph explanation method by extracting a Bayesian network from a sequence of graphs within an interval to serve as an explanation graph. HTGExplainer\cite{21} treats the dynamic graph sequence as a large static graph, i.e., multiple unrelated connected components, and directly applies static graph methods for explanation.

DyExplainer\cite{22} interprets dynamic graphs from two aspects through structural attention and temporal attention. STExplainer\cite{23} focuses on traffic network prediction and divides the model structure into temporal graphs and spatial graphs, utilizing graph attention networks to learn embeddings for both types of graphs to decouple spatiotemporal relationships. For task-state fMRI data classification, Ye Z\cite{24} et al. utilize perturbations on regions of interest (ROIs) in the brain network and calculate the importance of each ROI through the prediction gap after perturbation.

\subsection{Prototype explanations}

Prototype-based XAI can be categorized into two technical approaches: ante-hoc prototype-interpretable models and post-hoc prototype explanation methods \cite{111}. The former integrates prototype modules directly into network architectures, jointly optimizing predictive performance and prototype representations during training. In contrast, post-hoc prototype explanations identify or extract prototypes from pre-trained models without modifying their weights or requiring retraining. Research on ante-hoc methods is relatively extensive \cite{111, 112}, with notable examples including ProtoPNet\cite{101}, ProtoAttend\cite{103}, PIP-Net\cite{102}, and PCMNet\cite{107}.

Research on post-hoc prototype explanations remains relatively limited. Some studies have explored prototype extraction to interpret convolutional neural networks (CNNs) used for image classification. Examples include KMEx \cite{104}, its multi-layer variant \cite{105}, PCX \cite{106}, EPIC\cite{108}, and ProDG\cite{109}, which derive visual prototypes from fixed image models to explain classification outcomes. In the context of graphs, post-hoc prototype explanation techniques learn prototype subgraphs to interpret graph neural networks (GNNs) without modifying their parameters. For instance, Shin et al. \cite{113} develop graph prototypes to provide model-level explanations for GNN predictions. Although the post-hoc prototype methods offer flexibility for analyzing existing models, it is essential to ensure that the extracted prototypes genuinely reflect the model’s decision-making process rather than being artifacts of the extraction method. Furthermore, current graph prototype explainers focus exclusively on static graphs, leaving prototype-based explanations for dynamic graph models unexplored.

\section{Method}

\subsection{Problem definition}
In this paper, we introduce a prototype explainer for DGNN that identifies representative prototype examples of graph sequences to provide model-level explanations for the trained DGNN. An explanatory graph sequence is defined as a representative sequence of graphs that summarizes the key components of samples from a given class. Thus, in a DGNN model designed for a classification task, the explanatory graph sequence or prototype explanation \( S_c \) for a given class \( c \) satisfies by Formula \ref{eq:1}.

\begin{equation}
\min_{S_c} \sum_{i}^{n} dis(S_i, S_c) \tag{1}
\label{eq:1}
\end{equation}
Where \( S_i \) denotes the most significant subgraph sequence for the model within the \( i \)-th dynamic graph sequence of the dataset, and \( \textit{dis}(\cdot) \) is the function that measures the distance between two graph sequences. In this setting, a representative explanatory graph sequence for a specific class is determined by finding the graph sequence that has the smallest distance to the key subgraphs of all samples in that class.

To simplify the process of obtaining these representative explanatory graph sequences, we assume that these sequences are formed by ordering a series of representative graphs. These representative graphs, called prototype graphs, are extracted from the representative structures at each time point within the dynamic graph sequence. By summarizing all representative graphs present at each time point as prototype graphs and arranging these prototypes in a specific order for each class, we can derive the representative graph sequence—or prototype—that characterizes that class.

\subsection{Prototype explanation for dynamic graph neural networks}

\begin{figure}[htbp]
    \centering
    \includegraphics[width=0.5\textwidth]{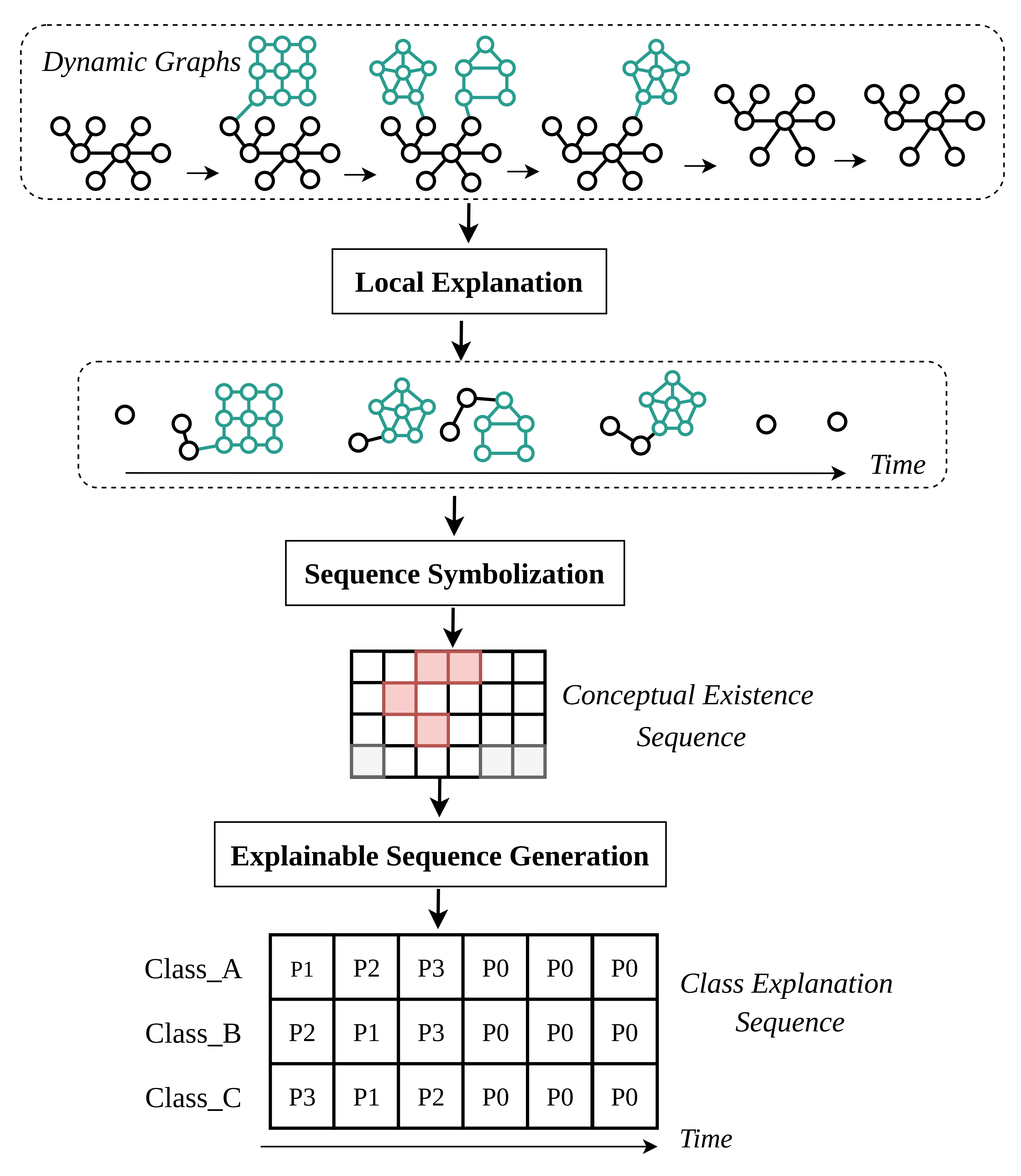}
    \caption{The Dynamic Graph Prototype Explainer follows a process that progresses from local to global explanations. Initially, it identifies significant subgraphs at each time step within dynamic graphs. Then, it generates representative prototype graph sequences by summarizing these subgraph structures.}
    \label{fig:2}
\end{figure}

A discrete dynamic graph neural network capable of learning graph structures can capture information about how a graph evolves over multiple time steps within a graph sequence. These structural changes are learnable and can be further abstracted and summarized. Accordingly, the DGPExplainer employs a two-step approach: first, it generates important subgraphs through local (instance-level) explanations; second, it constructs representative graph sequences using global (model-level) explanations to explain DGNN models, as illustrated in Figure \ref{fig:2}. 

To summarize the graph sequence and generate prototype explanations, the explainer first produces instance-level explanations at each time point by extracting the graph structures that the model focuses on at those moments. Next, the explainer applies a prototype-based, model-level explanation to summarize these graph structures: it extracts prototype graphs composed of these structures and constructs a sequence of prototype graphs that highlight the key features the model captures at each time point. Specifically, in this second step, the original discrete dynamic graph sequence is transformed into a standard symbolic sequence, and a sequence generation model is employed to learn and produce prototype graph sequences for each class, which serve as global explanatory examples for that class. In other words, by training generative models, all symbolic sequences within a class are learned and summarized to acquire global knowledge of the symbolic time series, thereby providing a comprehensive explanation.

\subsubsection{Instance-level explanation of dynamic graph neural networks}

In instance-level explanation, we extract the dynamic subgraph sequence that the model focuses on most for each sample by removing nodes and edges that are not critical to the model from the original dynamic graph sequence. The resulting explanatory graph sequence has the same length as the original dynamic graph sequence, and each graph at every time point is a subgraph of the corresponding snapshot.

As in common static graph explanation methods, our goal here is to select a subgraph sequence that is as small as possible while still producing prediction results as close as possible to those of the original sequence. That is, for an original dynamic graph sequence \( \hat{G}_i \), we aim to find a subgraph sequence \( G_i \) that satisfies Formula \ref{eq:3-2}, which extracts and outputs the most relevant subgraph by maximizing the mutual information between the explainable subgraph \( \hat{G}_i \) and the original model prediction \( \hat{y}_i \). 

\begin{equation}
\max_{\hat{G}_i} MI(\hat{y}_i, \hat{G}_i) = H(\hat{y}_i) - H(\hat{y}_i | G_i) \tag{2}
\label{eq:3-2}
\end{equation}
Where \( MI(\cdot, \cdot) \) denotes the mutual information, \( H(\cdot) \) denotes the information entropy, and \( H(\cdot | \cdot) \) denotes the conditional entropy. This optimization objective aligns with those used in many static graph instance-level explanation methods, allowing existing static graph explanation techniques to be slightly modified and applied to the explanation of dynamic graph neural networks.

However, directly applying static graph explanation methods to dynamic graph sequences can cause the model to lose focus on truly valuable time points. Moreover, providing the model with too many time points for unified optimization can hinder convergence due to the excessively large search space. In this paper, we introduce a two-step approach for instance-level explanations of dynamic graph neural networks: key time point extraction and structural subgraph extraction. The first step captures temporal information by identifying key time points, while the second step extracts the structural subgraphs corresponding to those time points to obtain spatial information.

\paragraph{Key time point extraction}

We employ explanation techniques of sequence models \cite{27,28,29} to determine the significance of each time point, retaining only those that exceed a predefined importance threshold. In implementation, for a given stacked DGNN model, the graph embedding layer computes the graph embedding at each time point based on the explanatory graph corresponding to that time point. For a sequence of explanation graphs \( G \), the graph embedding vector sequence \( E \), obtained through the calculation \( GNN(G) \), is masked with a mask \( M \) to produce \( \hat{E} \) during the explanation process. The calculation method is as follows:

\begin{equation}
\hat{E} = M \odot E + (1 - M) \odot \bar{E} \tag{3}
\label{eq:3-3}
\end{equation}

Where \( E \in \mathbb{R}^{d \times l} \), \( d \) represents the dimension of the graph embedding vector, \( l \) denotes the sequence length, and \( \bar{E} \) denotes the average graph embedding matrix obtained by expanding the mean vector \( \bar{e} \) of all sample graph embedding vectors to match the dimensions of the graph embedding sequence. The symbol \( \odot \) denotes element-wise multiplication.

This operation involves preserving the results obtained by passing significant time points through the explainer. For insignificant time points, the embedded graph is replaced with the average of all graph embeddings in the dataset to mitigate potential out-of-distribution (OOD) issues \cite{27}. Since thresholds may vary across different datasets, this method employs a dynamic threshold selection strategy, such as selecting the top-k edges or choosing an elbow point \cite{13}.

The attention-based method is a popular and effective approach for DGNN models \cite{2,3}. These models  inherently possess self-interpretability, enabling them to calculate the importance of each time step based on the input data. Therefore, for attention-based DGNN models, we can identify important time points by computing and comparing the attention weights within the DGNN as the input graph is processed.  

\paragraph{Structural subgraph extraction}
We use a static graph explainer for structural subgraph extraction to identify key subgraph structures within the DGNN model. The purpose of the static graph explainer is to output a subgraph derived from the original graph, as follows:

\begin{equation}
\hat{\mathcal{G}}_i = \text{Exp}(\mathcal{G}_i) \tag{4}
\end{equation}

Here, \( \mathcal{G}_i \) denotes the original graph, \( \hat{\mathcal{G}}_i \) denotes the explainable graph, and \( \text{Exp}(\cdot) \) denotes the explainer. To facilitate the adaptation of instance-level explanation methods from static graph models to dynamic graph models, the original dynamic graph sequence is concatenated into a large static graph in chronological order. Formally, this is expressed as \( \mathcal{G}_i = \text{CONCAT}_G(G_i) \), where \( \text{CONCAT}_G(\cdot) \) represents the graph concatenation process. The detailed procedure is presented in Algorithm \ref{alg:3-1}.

\begin{algorithm}
\caption{Dynamic graph merging}
\label{alg:3-1}
\begin{algorithmic}[1]
\Require Dynamic Graph Sequence \( G_i = [g_i^1, g_i^2, \cdots, g_i^l] \)
\Ensure Merged Static Graph \( \mathcal{G}_i \)
\State \( V \gets \{\} \) \Comment{Initialize the node set of the merged graph}
\State \( E \gets \{\} \) \Comment{Initialize the edge set of the merged graph}
\State \( \text{nodeOffset} \gets 0 \)
\For{$i$ from 1 to \( |G_i| \)} \Comment{Traverse all time points in the sequence}
    \State \text{currentNodes} $\gets g_i^t$.nodes()
    \State \text{currentEdges} $\gets g_i^t$.edges() \Comment{Obtain all nodes and edges on the graph at the given time point}
    \State \text{adjustedNodes} $\gets \{\}$ \Comment{The node set of the graph after initialization adjustment}
    \ForAll{node in \text{currentNodes}} \Comment{Renumber all nodes in the current graph}
        \State \text{adjustedNode} $\gets \text{node} + \text{nodeOffset}$
        \State \text{adjustedNodes} $\gets \{ \text{adjustNode} \} \cup \text{adjustNodes}$
    \EndFor
    \State \text{adjustedEdges} $\gets \{\}$ \Comment{The edge set of the graph after initialization and adjustment}
    \ForAll{edge in \text{currentEdges}}\Comment{Apply the same renumbering to all endpoints of the current edges}
        \State adjustedEdge $\gets$ 
        $(\text{edge.u} + \text{nodeOffset},\ \text{edge.v} + \text{nodeOffset})$
        \State $adjustedEdges \gets \{ adjustedEdge \} \cup adjustedEdges$ 
    \EndFor
    \State \( V \gets V \cup \text{adjustedNodes} \)
    \State \( E \gets E \cup \text{adjustedEdges} \) \Comment{Incorporate the adjusted nodes and edges into the merged static graph}
    \State \( \text{nodeOffset} \gets \text{nodeOffset} + |\text{currentNodes}| \) \Comment{Record the offset of the current graph from the subsequent graph}
\EndFor
\State \( \mathcal{G}_i \gets (V, E) \)
\end{algorithmic}
\end{algorithm}

As shown in Algorithm \ref{alg:3-1}, curly braces {} denote sets, while square brackets [] denote lists. The merge operation takes a sequence of dynamic graphs and returns a merged static graph. Steps 7 to 11 renumber each node in the graph at the current traversal step; steps 12 to 16 apply the same renumbering to the edges corresponding to these nodes. Finally, steps 17 and 18 merge the renumbered nodes and edges into the existing large static graph, which gradually grows as the time steps are traversed. Simultaneously, this process maintains a mapping from static graph nodes to the original dynamic graph nodes. Subsequently, this static graph can be used as input for the static graph instance-level explanation method to provide instance-level explanations of the static graph model. The explanation results are then converted back to the sequence of dynamic graphs using the static-to-dynamic graph mapping, thereby providing explanations for the DGNN model.

\subsubsection{Symbolization of dynamic graph sequences}

In this section, we present the method for using the above instance-level explanation graph sequences to symbolically represent dynamic graph sequences. By collecting instance-level explanations at each time point, as described above, we can construct a new dynamic graph dataset \( \hat{G} = \{\hat{G}_1, \hat{G}_2, \cdots, \hat{G}_n\} \), where each sequence \( \hat{G}_i = [\hat{g}_i^1, \hat{g}_i^2, \cdots, \hat{g}_i^l] \) consists of subgraphs explained at the sample level, corresponding to the original dynamic graph sequence. Next, we need to symbolize the graph at each time point; that is, map it to a specific literal value rather than to non-Euclidean spatial data represented by a graph.

We assume that the important graph sequence of each dynamic graph consists of several key substructures (motifs) arranged in a specific order. These key substructures are representative subgraphs at each time point and can therefore be referred to as prototype graphs. By identifying a set of prototype graphs and indicating which prototypes are present at each time point, we can characterize the significant graph structures throughout the entire sequence. Intuitively, by grouping these prototype graphs into a set and representing their presence at each time point with a binary vector of 0s and 1s, we capture the existing prototypes at each time point. We refer to this vector as the prototype existence vector. This approach provides a symbolic representation of the graph at each time point and, consequently, a symbolic representation of the dynamic graph sequence. 

To extract these important substructures, we propose a prototype classification model similar to those described in \cite{13,30}. This model also enables the extraction of the prototype existence vectors. This model also enables the extraction of the prototype existence vectors. By training this prototype classification model, the method learns an alternative model that faithfully reproduces the original model's output, thereby achieving a symbolic representation of the dynamic graph sequence.

In the graph sequence of instance-level explanation dynamic graphs, each time point typically consists of one or more connected components. We assume that each connected component corresponds to a single prototype graph; based on this assumption, the prototype classification model can be trained subsequently.

The prototype classification model utilizes a GNN to embed the connected components of each graph, producing an embedding vector \( e \) for the graph. First, a set of learnable prototype vectors \( P = \{p_i; i = 1 \cdots n_c\} \) is defined, where \( n_c \) denotes the total number of prototype vectors. Then, the graph embedding vector corresponding to each connected component is assigned to one of these prototype vectors using the Formula \ref{eq:3-8}.

\begin{equation}
Assign(e) = \text{softmax}\left(\frac{\text{sim}(p_1; e)}{\tau}; \ldots; \frac{\text{sim}(p_{n_c}; e)}{\tau}\right) \tag{5}
\label{eq:3-8}
\end{equation}

Where \( \tau \) is the temperature parameter; \( \text{sim}(\cdot, \cdot) \) is the similarity function used to calculate the similarity between two vectors, defined as follows:
\begin{equation}
sim(p_i; e) = \log\left(\frac{\left\|p_k - h\right\|_2^2 + 1}{\left\|p_k - h\right\|_2^2 + \epsilon}\right) \tag{6}
\label{eq:3-9}
\end{equation}

Where \( \epsilon \) is a small constant.Under this similarity measure, the similarity approaches infinity as the two vectors become closer and approaches zero as they move farther apart. Based on the calculated similarity, this method employs a temperature-adjusted softmax function to determine which prototype the connected component should be assigned to. In the process of calculating the assignment function, the parameter \( \tau \) serves as the temperature parameter. Using a temperature-adjusted softmax function allows for active control of the output, enabling it to be either smoother or more discrete. The lower the temperature, the greater the disparity in the softmax output values. In practical training, this method recommends gradually decreasing the temperature parameter as training progresses. This approach ensures smoother outputs in the early stages of training and promotes more discrete assignments by the prototype graph in later stages, effectively functioning like a hard assignment. Compared to the softmax with discretization used in GNNExplainer\cite{13}, this approach not only achieves faster convergence during the early stages of model training but also prevents information leakage to subsequent classification processes caused by non-zero weights, similar to the effect of discretization methods applied in later stages of training.

Based on the prototype assignment process described above, we calculate the prototype assignments for all connected components at each time point. Finally, by applying max pooling, we obtain the presence of all prototypes at that time point, represented by the prototype assignment vector \( a_i^t \) for the graph at that time point. This calculation is expressed by formula \ref{eq:3-10}.
\begin{equation}
a_i^t = \max(a_{i,1}^t, a_{i,2}^t, \cdots, a_{i,k}^t, \cdots, a_{i,n_k}^t) \tag{7}
\label{eq:3-10}
\end{equation}
Where \( a_{i,k}^t \) represents the prototype assignment vector of the \( k \)th connected component at time \( t \) in the \( i \)th sequence, and \( n_k \) is the total number of connected components at that time point. The `max` operation here takes the maximum value of the assignment for each feature—that is, the assignment of each prototype vector across all connected components. By arranging the prototype assignment vectors corresponding to each time-point graph in chronological order, we obtain the prototype assignment sequence \( A_i \) for the entire time series.

Finally, by inputting this prototype assignment sequence into the sequence embedding module, we obtain a sequence embedding vector. Feeding this vector into the classification head yields the model's classification result based on the prototype assignment sequence. It is essential to ensure that the final output of the prototype classification model closely approximates that of the original model, enabling it to mimic the original model's behavior. Therefore, we must optimize the objective function to achieve the optimal prototype classification model through this optimization process.

\begin{equation}
L_c(\theta) = \sum_{i=1}^{n} CrossEntropy\left(\psi_\theta(\hat{G}_i), \phi(G_i)\right) \tag{8}
\label{eq:3-11}
\end{equation}
\begin{equation}
\theta^* = \arg\min_{\theta} \sum_{i=1}^{n} CrossEntropy\left(\psi_{\theta}(\hat{G}_i), \phi(G_i)\right) \tag{9}
\label{eq:3-12}
\end{equation}
Meanwhile, referring to \cite{13}, we further introduce two additional constraints:
\begin{equation}
R_1 = \frac{1}{n_c} \sum_{k=1}^{n_c} \min_{\hat{G}_i} \left|\left| p_k - f_{GNN}(\hat{G}_i) \right|\right| \tag{10}
\label{eq:3-13}
\end{equation}
\begin{equation}
R_2 = \frac{1}{n} \sum_{i=0}^{n} \min_{k \in [1, n_c]} \left\| p_k - f_{GNN}(\hat{G}_i) \right\| \tag{11}
\label{eq:3-14}
\end{equation}
Here, \( f_{GNN}(\cdot) \) denotes the function used to compute graph embeddings. The first constraint ensures that each prototype is close to at least one sample, while the second ensures that each sample is relatively close to a specific prototype. The loss function is then defined as follows:

\begin{equation}
L(\theta) = L_c(\theta) + \lambda_1 R_1 + \lambda_2 R_2 \tag{12}
\label{eq:3-15}
\end{equation}
Here, \( \theta \) represents the parameter set used in the prototype classification model, encompassing all trained prototype vectors. The function \( \psi_{\theta}(\cdot) \) denotes the prototype classification model parameterized by \( \theta \), which includes a graph embedding module, a prototype assignment module, a sequence embedding module, and a classification module. The terms \( \lambda_1 \) and \( \lambda_2 \) are adjustable hyperparameters. 

By minimizing the final loss, we train a prototype classification model that closely approximates the output of the original model, thereby obtaining its corresponding optimal parameter \( \theta^* \). Subsequently, based on the obtained optimal parameters, we collect the optimal tokenized sequences \( A^* \) corresponding to all sequence sets, represented as \( A^* = \text{Assign}_{\theta^*} \left( GNN_{\theta^*}(\hat{G}_i) \right) \), where \( \text{Assign}_{\theta^*}(\cdot) \) denotes the prototype assignment process using the optimal parameter \( \theta^* \), and \( GNN_{\theta^*}(\cdot) \) represents the graph embedding process employing the optimal parameters.

\subsubsection{Generation of explanation graph sequences}

Based on the sequence symbolization process described above, we obtain the prototype existence sequence corresponding to each graph sequence. To generate explanation graph sequences, we discretize the prototype existence sequence using a threshold of 0.5. We then aggregate the prototype existence sequences for each class to produce a graph sequence that serves as the global explanation for that class. Specifically, we employ a sequence generation model to learn the pattern of each sequence through self-supervised learning and subsequently generate a graph sequence as the global explanation for each class. Additionally, since a sequence pattern may occur at any point within the sequence, we apply methods such as elbow selection or a fixed threshold \cite{13} to the importance scores of time points obtained during the instance-level explanation process. This allows a starting time point to be selected for the sequence, which is then provided as input to the sequence model.

Specifically, during generation, the sequence model takes as input the class information, the starting time point of the sequence, and the previously generated sequence data to produce the next time point, thereby fitting the entire sequence. This approach is common in autoregressive tasks such as text generation. However, since multiple prototypes may be present at each time point, the vocabulary size does not directly correspond to the number of prototypes. Otherwise, the vocabulary size would be on the order of \( vocabulary\_size = 2^{n_c} \), which becomes impractical as the number of prototypes increases. To address this, inspired by the Byte Pair Encoding (BPE) method \cite{31} in natural language processing, we flatten the concept existence sequence when generating a complete prototype existence sequence and sequentially generate the probability of each concept's presence at each time point. This reduces the vocabulary size to 2, with each decision determining whether the current position is 0 or 1. 

Consequently, the objective function for training this sequence model can be expressed as follows:

\begin{equation}
L(\theta) = - \sum_{j=1}^{T} \log P \left( b_j \mid b_{<j}, x; \theta \right) \tag{13}
\label{eq:3-16}
\end{equation}
That is, we seek a model that maximizes the probability of the current \( b_j \) given the preceding values \( b_{<j} \), where \( b_j \) represents the value at the j-\text{th} position in the flattened prototype existence sequence, which is either 0 or 1.

In practice, the model consists of two components: a sequence module and a classification head. The sequence module generates hidden layer vectors based on the class information, the starting time point, and the past sequence. The classification head then determines the value at each position in the sequence using these hidden vectors. The formula for the sequence model is as follows:

\begin{equation}
h_j = f_{\text{seq}}(c_t, h_{j-1}) \tag{14}
\label{eq:3-17}
\end{equation}
Based on the hidden state, a classification head is used to compute the probability that the output at this time point is 0 or 1:
\begin{equation}
b_t = \text{sigmoid}\left(f_{\text{cls}_{\text{head}}}(h_{t-1})\right) \tag{15}
\label{eq:3-18}
\end{equation}
Where  \( h_0 \) is calculated using other methods that incorporate class information and initial point data.

After completing model training, we set the starting position to 0 and the class to the target class to obtain a complete explanation sequence. During actual generation, the model first randomly generates a specified number of prototype existence sequences based on the probabilities output by the generation model. Next, for each generated prototype existence sequence, the model selects the prototype sample with the closest distance on each prototype vector as the graph at that time point. Finally, using feedback from the model, it selects the prototype sequence whose prediction result most closely matches the specified class as the explanation sequence instance for that class.

\section{Experiment}

\subsection{Dataset description}

We use both synthetic and real-world datasets to evaluate the proposed TPGExplainer for the task of explanation. 

(1) Synthetic dataset

\begin{figure}[htbp]
    \centering
    \includegraphics[width=0.5\textwidth]{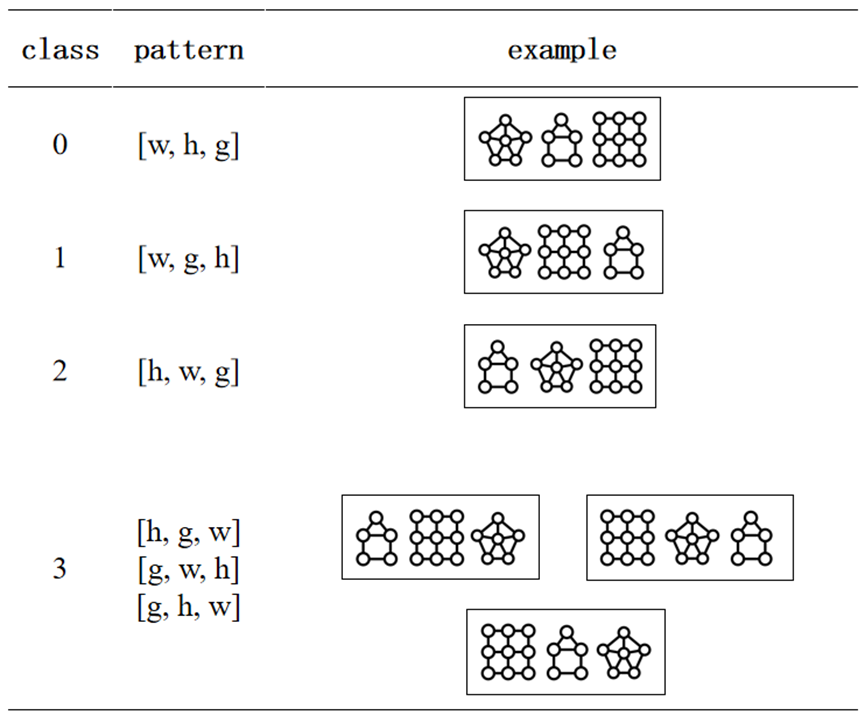}
    \caption{Description of the Dyn-BAMotif dataset, and its pattern graphs: w represents a wheel, h represents a house, and g represents a 3x3 grid. }
    \label{fig:4-1}
\end{figure}

Referring to the BA-2Motif dataset used in GNNExplainer\cite{32}, we design a synthetic dataset, Dyn-BAMotif, specifically for DGNN explanation. The dataset contains 1,000 samples. To maintain contextual coherence between graphs at different time points, the graphs in each sequence share the same base graph, which is a randomly generated scale-free network. In practice, all base graphs are generated using the Barabási-Albert model, with each base graph consisting of 10 nodes. After generating the sequence of base graphs, the class to which each sequence belongs is randomly assigned. Then, a pattern sequence matching the class is randomly selected and incorporated into the base graph sequence at a random time point. This dataset includes all permutations of three pattern graphs—house, 5-wheel, and 3×3 grid—as the pattern sequences that can appear in the classes. As shown in Figure \ref{fig:4-1}, Dyn-BAMotif places a total of six permutations of the three pattern graphs into four classes. The class labeled 3 contains three different pattern sequences, which are evenly distributed among all data labeled as class 3.

(2) Real-world datasets

Two real-world dataset were utilized in the experiment. First, the Autism Brain Imaging Data Exchange I (ABIDE-I) \cite{33} is a comprehensive collection of resting-state functional magnetic resonance imaging (rs-fMRI) data from patients with autism spectrum disorder (ASD) and healthy controls (HC). This dataset includes data from 17 international sites, providing a diverse and extensive resource for autism research. Previous studies have shown that classification accuracy is higher when analyzing scan data from a single site but significantly decreases when analyzing data across multiple sites, likely due to differences in scanning parameters. Following \cite{2}, data from the ABIDE I Preprocessed Connectome Project \cite{34} were also used in the experiments. These data were preprocessed using the Configurable Pipeline for the Analysis of Connectome (CPAC), subjected to bandpass filtering, and no global signal regression was performed. The dataset comprises a total of 871 subjects. During the experiment, all subjects with a scan duration of less than 160 seconds were excluded, resulting in a final dataset of 667 samples, including 306 ASD patients and 361 healthy controls. This dataset provides a crucial foundation for advancing the understanding and detection of autism through innovative graph-based learning techniques. In our experiments, functional connections (FC) within each window were constructed from the collected fMRI signal sequences using a sliding window method, resulting in a dynamic functional connectivity (DFC) sequence. Subsequently, the functional connectivity matrix at each time point was thresholded, retaining values above the threshold as the adjacency matrix for the graph at that time point. The rows of the original functional connectivity matrix were used as node features to construct dynamic graph data. All sequences were uniformly truncated to a length of 160 in the experiment, with data from the first 10 time points discarded. A sliding window of size 50 and step size 10 was used to calculate dynamic functional connectivity, and the top 10\% of edges at each time point were selected as the graph edges for classification.

Second, the dblp-ct1 dataset \cite{35} is a subset of the DBLP database used to generate temporal co-author graphs and initially includes 60 sampled nodes. This dynamic graph dataset includes two types of labels for categorizing different dissemination processes. Interactions between these nodes are recorded at regular time intervals, resulting in a sequence of discrete dynamic graphs with a fixed length of 48 time points. An edge is created between two nodes at a given time point if an interaction occurs between them during that interval. A total of 755 distinct dynamic graph sequences were generated through 755 samplings. Subsequently, two different propagation models were applied to these dynamic graph sequences. One model employs random propagation, where infected individuals are randomly added at each time point. The other model uses a susceptible-infected (SI) propagation model with an infection probability of 0.5. The dynamic graph sequences generated by these two propagation methods are treated as different classes.


\subsection{Evaluation metric}

We use three indicators—fidelity, hit rate, and completeness—to evaluate the explanation methods.

\textbf{Fidelity}: Fidelity measures whether the explanations provided by the model accurately reflect the model's behavior. At the model-level explanation, it refers to the explanation given for a specific class, and Fidelity assesses whether the model produces the expected classification result for the explained class. Let the complete set of explanations for each class be \( \Omega \). Fidelity can be calculated using formula \ref{eq:4-1}:

\begin{equation}
Fidelity(\Omega) = \frac{1}{N_c} \sum_{i=0}^{c} Eq(\Phi(\Omega_i), i) \tag{16}
\label{eq:4-1}
\end{equation}
Where \( N_c \) represents the total number of classes, \( \Omega_i \) denotes the explanation graph corresponding to class \( i \) in the explanation set, and \( Eq(\cdot, \cdot) \) is a function used to determine whether two results are consistent.

\textbf{Hit rate}: For datasets such as Dyn-BAMotif, which contain factual graph sequences, the relationship between the explanation results and the factual graph sequences can be leveraged to evaluate the quality of the explanations. The hit rate is computed by collecting all explanation graph sequences in the explanation set, verifying their consistency with the factual graph sequences, and calculating the proportion of explanation graphs that appear in the factual graph. Let the set of explanation graphs that are present in the factual graph (True Positives) be denoted as \( \Omega_{TP} \), and the set of explanation graphs that are faithful to the model (Positives) be denoted as \( \Omega_P \). The hit rate is then calculated using formula \ref{eq:4-2}:

\begin{equation}
Hitness(\Omega) = \frac{\left|\Omega_{TP}\right|}{\left|\Omega_P\right|}
\tag{17}
\label{eq:4-2}
\end{equation}

\textbf{Completeness}: Completeness measures whether the explanation method can identify a complete sequence of fact graphs in the dataset. Let the set of fact graph sequences (Ground Truth) be denoted as \( \Omega_{GT} \). Then, the completeness of the explanation result set can be calculated using formula \ref{eq:4-3}:

\begin{equation}
Completeness(\Omega) = \frac{\left| unique(\left( \Omega_{TP} \right) \right|}{\left| \Omega_{GT} \right|} \tag{18}
\label{eq:4-3}
\end{equation}

In practical calculations, the hit rate and completeness are determined by considering cases where only some subgraphs in the sequence are consistent. For example, in a subgraph sequence of length 3, if only one time point is consistent, then the hit rate for this subgraph sequence is \( \frac{1}{3} \).

\textbf{Average explanation graph size}: The average explanation graph size (AveSize) quantifies the number of edges that require explanation when the method provides reliable explanations. Let \( E_i^j \) denote the set of edges in the graph at the \( j \)th time point of the \( i \)th dynamic graph sequence in the explanation results. Then, the average explanation graph size is defined as:

\begin{equation}
AveSize = \frac{1}{|\Omega|}{|\sum_{i=0}^{|\Omega|} \frac{1}{L_i} \sum_{j=0}^{L_i} \left|E_i^j\right|} \tag{19}
\label{eq:4-4}
\end{equation}
Where \( \Omega \) denotes the size of the set of explanation results mentioned above, and  \( L_i \) denotes the length of the \( i \)th explanation sequence. By summing the number of edges contained in all dynamic graph sequences within the explanation results and then calculating the average, we obtain the average size of the graphs in the explanation results. This metric allows us to assess the conciseness of the explanation results and, to some extent, indicates their comprehensibility.

\subsection{Experimental environment and the model being explained }

All experiments in this paper were conducted using Python version 3.12.8. Implementations related to neural networks and graph neural networks utilized PyTorch version 2.5.1 and its associated graph neural network library, PyTorch Geometric version 2.6.1, respectively.

We introduce the DGNN model for explaining, which features a stacked architecture consisting of a graph embedding module, followed by a sequence embedding module, and concluding with a classification head for final classification. The DGNN model was configured and trained using the ABIDE-I, Dyn-BAMotif, and dblp-ct1 datasets. 

\textbf{Dyn-BAMotif}. The DGNN model applied to the Dyn-BAMotif dataset features a three-layer architecture, with each hidden layer having a dimensionality of 32. Each GNN layer comprises a stack of two linear layers. The output of each linear layer is first passed through a Batch Normalization (BatchNorm) layer, followed by an activation function using Leaky ReLU with an \( \alpha \) value of 0.01. After obtaining the embedding vectors for each node in the graph at each time point via the GNN layers, global max pooling is applied to compute the graph embedding at each time point. The embeddings of the time-point graphs are then organized into a sequence, and positional encoding is computed following the method described in \cite{11}, utilizing both cosine and sine functions. The sequence embedding module employs a multi-head self-attention mechanism, which processes the graph embedding sequence augmented with a cls\_token and positional encoding. Prior to the self-attention module, a Dropout layer with a probability of 0.1 is applied. The multi-head attention module uses four heads. The model is trained using the Adam optimizer with a learning rate of 0.01 over 2000 epochs.

\textbf{ABIDE-I}.The overall architecture of the predicted dynamic graph neural network applied to the ABIDE-I dataset is consistent with that used in Dyn-BAMotif. However, due to the large graph scale, the GNN is modified to a six-layer GCN model. A Dropout layer with p = 0.5 is added after each GCN layer. The pooling layer for graph embedding is changed to incorporate both global max pooling and global average pooling, with the outputs of these two pooling layers concatenated to form the graph embedding. The sequence embedding module also employs self-attention, but a linear layer is added before it, with LayerNorm layers applied both before and after the linear layer. Additionally, the Dropout layer preceding the sequence embedding layer is set to p = 0.5. The Adam optimizer is used with a learning rate of 0.0001, and the model is trained for 500 epochs.

\textbf{dblp-ct1}. The settings used for the dblp-ct1 dataset were similar to those of Dyn-BAMotif; however, in this experiment, the model's hidden layer dimension was set to 16, and only two layers of GIN were employed. The number of training epochs was set to 50, and the learning rate was 0.001.

The performance of the DGNN on these datasets is shown in Table \ref{tab:4-1}.

\begin{table}[htbp]
\centering
\caption{Performance of the DGNN model across different datasets (\%)}
\label{tab:4-1}
\begin{tabular}{lccc}
\hline
Dataset & Accuracy & ROCAUC & F1-Score \\
\hline
Dyn-BAMotif & 96.00 & 99.75 & 95.333 \\
dblp-ct1 & 96.70 & 99.60 & 96.70 \\
ABIDE-I & 71.09 & 77.25 & 65.42 \\
\hline
\end{tabular}
\end{table}

\subsection{Comparison models and settings}
Due to the lack of directly applicable comparison schemes for DGNN prototype explanations, we adapted two existing methods from model-level explanations of static GNN models to serve as comparison schemes.

\subsubsection{XGNN+}
We extended XGNN\cite{25} to XGNN+ to explain the DGNN model using prototypes. Originally, XGNN employed reinforcement learning to generate explanations for graph neural networks. In this paper, we modified the generation process to produce dynamic graphs, resulting in XGNN+. Similar to extending local explanation methods from static to dynamic graphs,  we concatenate the explanatory images corresponding to each snapshot in the dynamic graph to generate a large graph. We can then select any node in this large graph as the starting node. However, when predicting the ending node, the selection mask is restricted: instead of allowing all existing nodes and candidate set nodes, it only permits nodes present in the graph at the same time point and nodes in the candidate set. This restriction prevents the generation of edges between different time points. Consequently, this approach produces a dynamic graph sequence composed of multiple distinct graphs. For the Dyn-BAMotif dataset, considering its characteristics, the sequence length is set to 3 in the experiments, with 15 nodes at each time point.

\subsubsection{GNNInterpreter+}
GNNInterpreter \cite{26} provides model-level explanations for each class by employing a gradient-based approach to train a sampler that maximizes the model's output for the corresponding class. In this paper, we extend GNNInterpreter to GNNInterpreter+ by adapting it for the dynamic graph domain. We divide the parameters used during sampling into multiple vectors—equal in number to the sequence length—and optimize these vectors simultaneously, enabling the sampler to generate multiple graphs at once. Arranging these graphs sequentially produces an explanatory graph sequence. In experiments on Dyn-BAMotif, the generated graph sequence has a length of three, with ten nodes at each time point.

\subsubsection{TPGExplainer}
The TPGExplainer proposed in this paper requires extracting local explanations for dynamic graphs using local explanation methods. In the experiment, PGExplainer, which employs static graph explanation methods, is extended to extract local explanations for dynamic graphs. For the parameter settings of PGExplainer, the constraint on the size of the explanation graph is set to 1e-7, and the constraint on the discretization degree of the explanation graph is set to 0.1. The optimizer used is Adam, with a learning rate of 0.001. On the Dyn-BAMotif dataset, after obtaining the edge mask for local explanations, the elbow points in the mask are calculated to determine thresholds. On the ABIDE-I dataset, to maintain edge sparsity, the top 5\% of edges with the largest mask values (indicating higher importance) are retained in the experiment.

In the prototype classification model, the prototype vectors are initialized randomly. During training, the temperature parameter decreases exponentially with a base of 0.98, raised to the power corresponding to the number of training epochs. The optimizer used is Adam. The learning rates vary between components: the GNN and prototype vectors have a learning rate of 0.001, while the sequence model and classification module use a learning rate of 0.01. This configuration ensures that the prototype vectors remain as stable as possible while allowing the classification module to quickly adapt to changes in the prototype vectors. The prototype classification model is trained for 500 epochs on both datasets, with the temperature gradually decreasing from 0.98 to approximately 0.00000004. On the Dyn-AMotif dataset, six prototype vectors are selected, with \( \lambda_1 \) set to 0.09 and \( \lambda_2 \) set to 0.00099; on the ABIDE-I dataset, eight prototypes are used, with \( \lambda_1 \) set to 0.1 and \( \lambda_2 \) set to 0.003. On dblp-ct1, five prototype vectors are selected, and the remaining settings are consistent with those used for Dyn-AMotif.

\subsection{Experiment and result analysis}

\subsubsection{Model explanation experiments on synthetic dataset}

For each method, we ran it 10 times using different random seeds and then calculated the average of all results. The obtained results are presented in Table \ref{tab:4-2}.

\begin{table}[htbp]
\centering
\caption{Performance of prototype explanation of DGNN models (mean±std)}
\label{tab:4-2}
 \begin{tabular}{lcccc}
\hline
Method & Fidelity (\%) & Hitness (\%) & Complete. (\%) & AveSize \\
\hline
XGNN+ & 46.68±41.85 & 14.17±6.50 & 9.44±4.33 & 9.33±0.49 \\
GNNInterpreter+ & 78.71±12.14 & 27.78±1.62 & 18.51±0.72 & 35.47±4.35 \\
TPGExplainer & \textbf{93.34±13.31} & \textbf{93.34±13.31} & \textbf{62.24±8.92} & \textbf{11.67±0.00} \\
\hline
\end{tabular}
\end{table}

To reflect a certain degree of explanatory power for each method, even if the method cannot provide a complete sequence of facts, the corresponding indicators are calculated based on the portion of the graph that contains accurate facts. For example, for the pattern [w, h, g], if only w is accurate and the other two elements are inaccurate, the number of graphs correctly identified by the sequence is recorded as 1. If the sequence is completely accurate, the number of graphs hit is 3. Additionally, since GNNInterpreter+ is derived by removing edges from a complete graph, when the number of edges in the graph exceeds 30—which is two-thirds of the complete graph—the graph is recorded as a miss.

Based on the experimental results above, TPGExplainer demonstrates the best performance among the methods evaluated. Next, we will conduct a visual analysis of the explanation results for TPGExplainer and XGNN+. The analysis of TGPE-base results will be presented later in the ablation experiments.

\begin{figure}[htbp]
    \centering
    \includegraphics[width=0.7\textwidth]{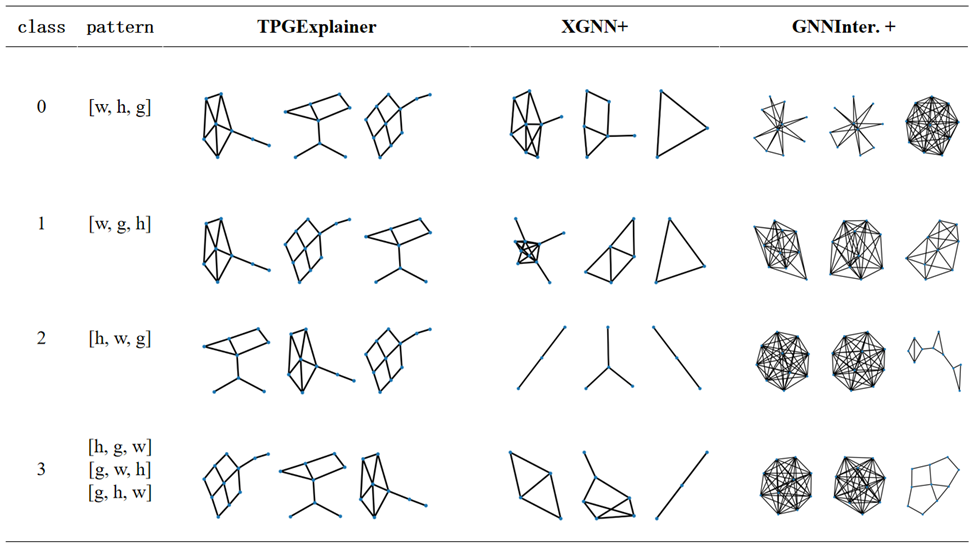}
    \caption{Comparison of generated model-level explanations}
    \label{fig:4-2}
\end{figure}

The explanation results shown in Figure \ref{fig:4-2} indicate that XGNN selects explanations closer to real samples within the generated results. However, as illustrated in the figure, compared to XGNN+, TPGExplainer produces explanations that are more interpretable and better aligned with the dataset. Although the explanations generated by XGNN+ ensure consistent model behavior, they significantly deviate from the actual graph structures present in the dataset. The explanations for Class 0 are relatively accurate, with the first two graphs containing wheel and house substructures. Conversely, the explanations for Class 2 are the poorest, ultimately converging to an empty graph because the model assigns the highest classification probability to Class 2 when given an empty graph. This demonstrates that during the generation of explanation graphs, XGNN, lacking mechanisms to mitigate out-of-distribution (OOD) issues, produces many samples outside the distribution learned by the model. In contrast, TPGExplainer ensures that all prototype graphs originate from instance-level explanations of the original dataset, guaranteeing that each time-point subgraph in the explanations remains within the original data distribution.

\subsubsection{Model explanation experiments on brain network dataset}

\begin{figure}[htbp]
    \centering
    \includegraphics[width=0.95\textwidth]{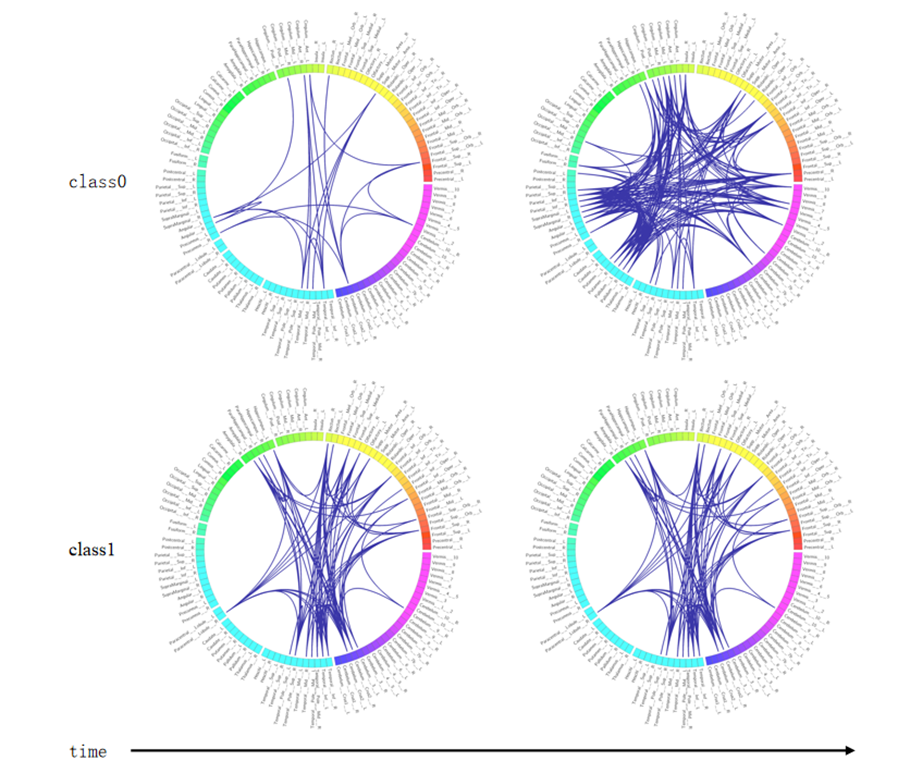}
    \caption{explanation of PTGExplainer on brain network data}
    \label{fig:4-3}
\end{figure}

For the model explanation experiment on brain network data, due to the long complete time series, an instance was selected from each class, and a subsequence of length 2 within that instance was analyzed. The prototype sample at each time point was determined by selecting the sample from the existing prototypes described by the prototype existence vector that is closest to the corresponding prototype vector. The visualization results are shown in Figure \ref{fig:4-3}, where class 0 represents the control group and class 1 represents the ASD group.

From the visualized results, it can be observed that the brain connectivity related to the posterior cingulate cortex in the ASD group is significantly weakened, which is consistent with the findings of some previous studies \cite{37,38}. Furthermore, in the ASD group, the connectivity related to the right angular gyrus has also weakened, which aligns with past research results \cite{39}.

\subsubsection{Model explanation experiments on co-author dataset}

\begin{figure}[htbp]
    \centering
    \includegraphics[width=0.4\textwidth]{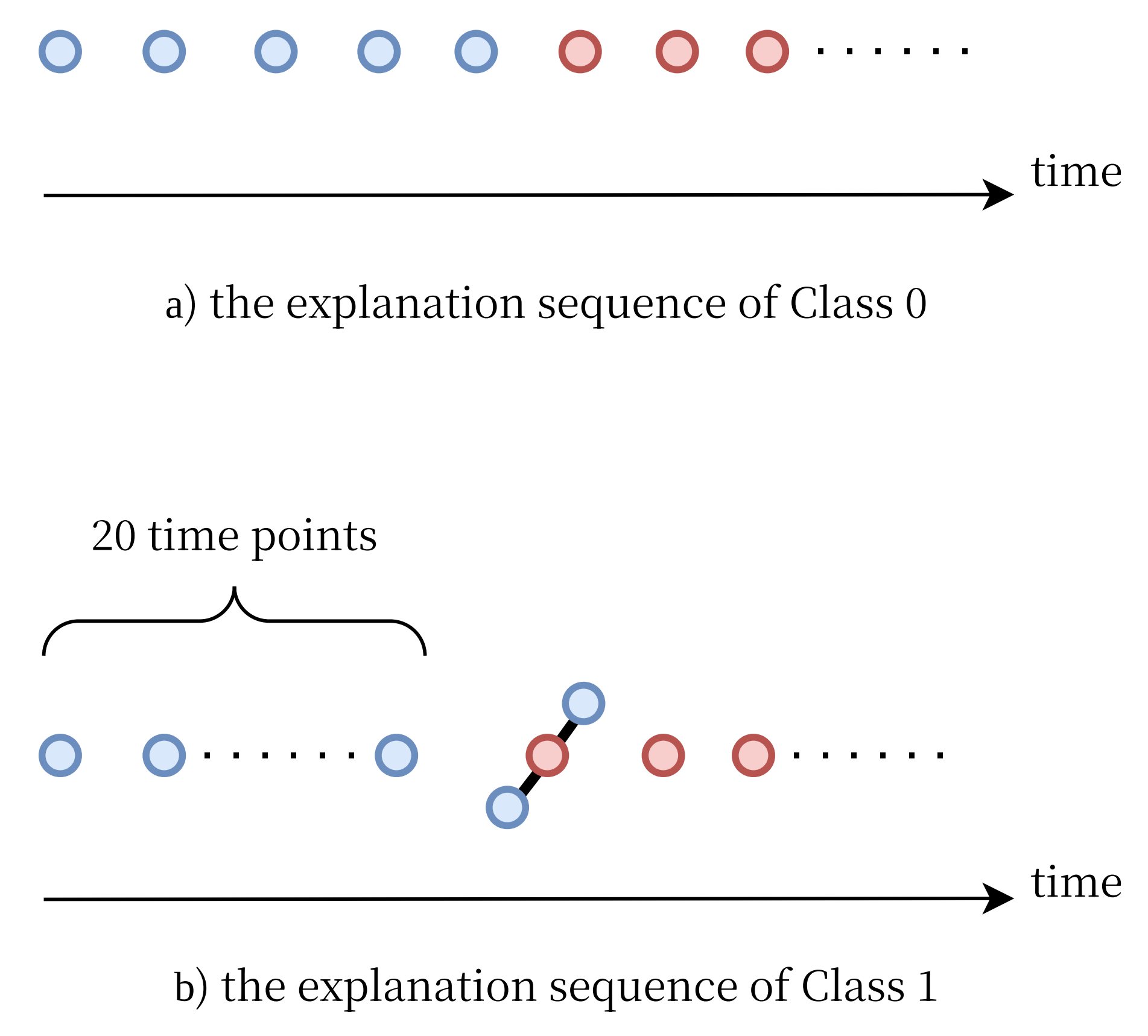}
    \caption{Experimental results of model explanation by TPGExplainer on communication data}
    \label{fig:4-4}
\end{figure}

In the dblp-ct1 dataset experiment, a representative explanation sequence was selected for each of the two classes, illustrating distinct patterns in the different dissemination processes of the co-author networks. The explanation sequences for these classes are as follows. In the figure, blue nodes represent uninfected nodes, while red nodes represent infected nodes. class 0 corresponds to the sequence obtained through random propagation, whereas class 1 corresponds to the sequence derived from the SI propagation model. Since multiple nodes can appear at each time point, these nodes form several connected components. Additionally, the sequence typically ends with both uninfected and infected nodes, each potentially belonging to different connected components. Each connected component is assigned to a prototype. To simplify the presentation of experimental results, the figure displays the closest graph to a representative prototype at each time point. As shown in Figure \ref{fig:4-4}b), the first 20 time points in the class 1 sequence consist of graphs containing only one blue node. This indicates that, at these time points, all nodes in the original graph are uninfected, and no edges remain in the local explanation results. Consequently, there is only a prototype consisting of uninfected nodes. The last few time points show graphs containing only one red node, meaning that at these times, both uninfected and infected nodes are present, but the figure only displays the prototype represented by the infected nodes.

From the results shown in the figure, it can be observed that class 0 has infected nodes appearing at earlier time points, whereas class 1 experiences its first infected node only after a longer delay. At the onset of infection, there is typically a connection between the initial infected node and other uninfected nodes. This sequence of events aligns with the dataset's definition: in random contagion, infected nodes emerge earlier when there are no edges in the dynamic graph, while in the dynamic graph sequence generated using the SI propagation model, the first infected node appears only after a connection between two nodes is established, followed by the emergence of independent infected nodes as prototypes.

\section{Conclusion}
This paper focuses on researching model-level explanations for DGNNs, primarily by deriving explanatory instances from prototype graph permutations. We propose TPGExplainer, a prototype-based model-level explanation method for DGNNs. This approach extracts prototype graphs from the input graph at each time point and organizes these prototypes to generate representative examples for each class, serving as model-level explanations. It extends an improved static graph instance-level explanation method to dynamic graph sequences. Based on the instance-level explanation results, the method decomposes the graph at each time point into distinct prototypes, thereby symbolizing the dynamic graph sequence. Finally, using the symbolized sequence, a sequence generation technique produces an explanatory dynamic graph sequence corresponding to each class.

In summary, the prototype-based model-level explanation method proposed in this paper provides an effective solution for explaining DGNN models by transforming their complex decision-making processes from abstract mathematical operations into human-understandable, analogical, case-based explanations. In other words, when the model classifies an input into a particular class, it compares that input to a representative example—a prototype graph sequence identified by the TPGExplainer. In the future, we will focus on the following aspects: thoroughly investigating the generalization capabilities of current methods and applying them to a broader range of real-world datasets, as well as developing more general models to expand the applicability of explanation methods, which are currently primarily designed and tested for stacked DGNNs.


\bibliographystyle{unsrt}  
\bibliography{references}

\end{document}